# A SURVEY OF MACHINE LEARNING ALGORITHMS FOR DETECTING RANSOMWARE ENCRYPTION ACTIVITY

Erik Larsen, David Noever, Korey MacVittie

PeopleTec, Inc., 4901 Corporate Dr., Huntsville, AL 35805
erik.larsen@peopletec.com

## ABSTRACT

*A survey of machine learning techniques trained to detect ransomware is presented. This work builds upon the efforts of Taylor et al. in using sensor-based methods that utilize data collected from built-in instruments like CPU power and temperature monitors to identify encryption activity. Exploratory data analysis (EDA) shows the features most useful from this simulated data are clock speed, temperature, and CPU load. These features are used in training multiple algorithms to determine an optimal detection approach. Performance is evaluated with accuracy, F1 score, and false-negative rate metrics. The Multilayer Perceptron with three hidden layers achieves scores of 97% in accuracy and F1 and robust data preparation. A random forest model produces scores of 93% accuracy and 92% F1, showing that sensor-based detection is currently a viable option to detect even zero-day ransomware attacks before the code fully executes.*

## 1. INTRODUCTION

The growing threat of ransomware has escalated with the development of advanced security evasion techniques such as metamorphic code that can alter itself after each execution. This renders signature-based detection algorithms useless, and zero-day deployments of new ransomware even harder to detect. Building upon the work of Taylor et al. [1] and Marais et al. [2] this report details the findings from training a diverse set of machine learning models on side-channel data gathered from the computer hardware. Detecting file encryption with data from sensors monitoring battery voltage, clock speed, temperature, and other physical attributes is shown to be a viable approach in [3-5].

### 1.1. Background

Performance monitoring tools collect information from sensors mounted on computer hardware. These systems typically run by default and are used to track the health and operating characteristics of the machine's physical and system resources. Metrics of interest – e.g., CPU core temperature, CPU load, % idle time, clock readings, and memory usage – can be gathered from the CPU, disk, network, and memory (physical) or even handles and modules (system). Collected data can be configured, scheduled, and reviewed either in real time or from stored performance logs. When a computer is given a task, such as file encryption, the load on the system changes. Consequently, relevant sensors continuously detect this physical activity then report to any number of collectors which log the data and can initiate appropriate responses. For example, when a core temperature is too hot, the cooling fan is switched on. Relating to security, real time monitoring of collected data enables machine learning algorithms, trained in identifying malicious activity, to alert when a cyber-attack occurs, preventing its completion.

## 2. DATA PREPARATION

Data used to train the models was obtained by tracking internal sensor outputs as files were encrypted, simulating both malicious and normal activity. A human labeled dataset consisting of 47 features identifies this activity as encryption being "OFF" or "ON" – encoded as 0 and 1 – making it a binary supervised learning problem. As shown in Figure 1, there are 26,728 examples: 16,549 type 0 and 10,179 type 1. Variables recorded include CPU and GPU loads, hard disk drive (HDD) and CPU temperatures, RAM load, and more [5]. The test harness includes OpenHardwareMonitor [6] for logging performance metrics across multiple (4) CPUs and hard drive write functions.

As shown in Figure 2, each featured category has its statistical value regime. Clock_0 shows no standard deviation, while temperature_4 has a wider spread of numbers across quartiles. Distributions show that these fifteen categories use the same values. There is also some separation in distributions between the two targets, with readings being higher for the "ON" situation as seen in Figure 3.

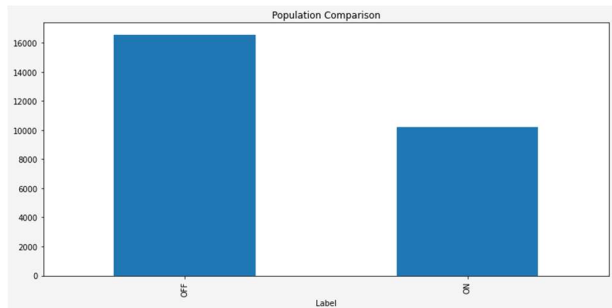

Figure 1. Class sizes

Several columns contain duplicate information. One of each of these pairs can be dropped from the original data. CPU temperatures, loads, and clock speeds appear to be the most influential and were thus selected to simplify the model. Categorical variables are encoded into vectors using the pandas get_dummies function. The final format is saved in a CSV file with 15 feature columns all of type int64 or float64.

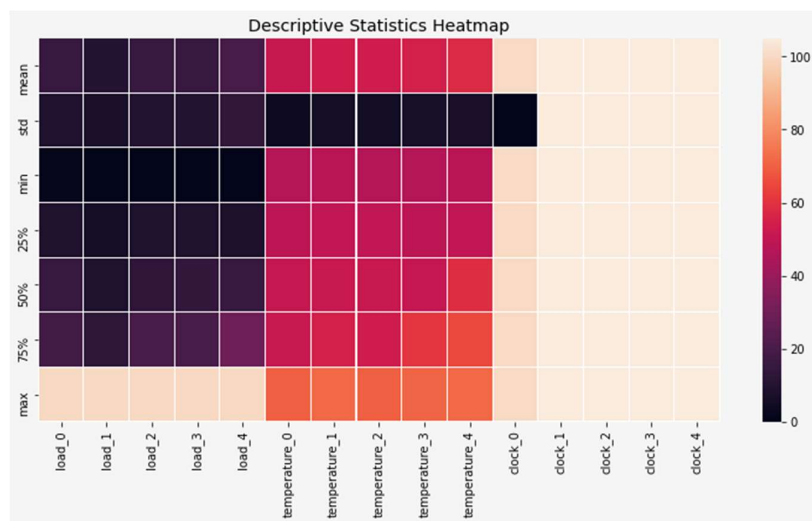

Figure 2. Differing value profiles for each feature type

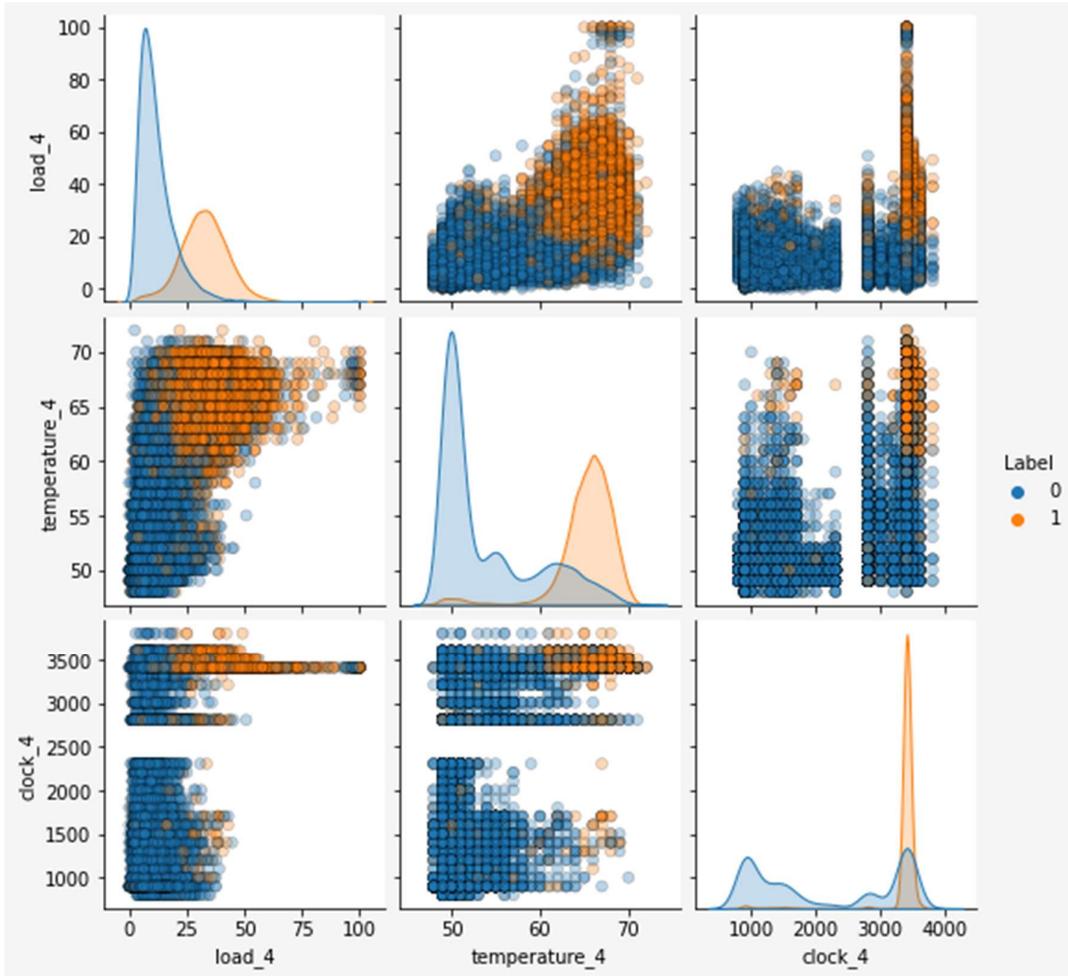

Figure 3. Separation is seen in the distributions for the "4" processor features

## 3. Model Comparisons

The PyCaret module [7] is first used to compare several models at once. Multiple comparisons are made using differing numbers of the selected features mentioned above in various combinations. The top three performers are identified, and further optimization using Scikit Learn's libraries and a comprehensive grid search with three-fold cross-validation are carried out to maximize performance with as little data manipulation as possible. This is done with the idea of keeping processing time to a minimum in order to react before a zero-day attack is completed. Other models such as sequential dense neural networks and Multilayer Perceptron are explored as well. When considering the extension to enterprise production, it is recommended to use at least 5-fold cross-validation.

### 3.1 PyCaret

Thirteen models are trained, scored, and listed by rank for the entire complement of 15 features selected from the ransom_sim dataset (see Table 1). A second comparison (Table 2) shows a performance loss in all metrics when using only three of these feature columns, though computation time is reduced by roughly 30%.

Table 1. PyCaret Model Comparison 1 Results.

| Model | Accuracy | AUC | F1 | MCC | TT (Sec) |
|---|---|---|---|---|---|
| Random Forest Classifier | 0.9281 | 0.9660 | 0.9064 | 0.8482 | 3.468 |
| Extra Trees Classifier | 0.9279 | 0.9659 | 0.9064 | 0.8480 | 3.194 |
| Light Gradient Boosting Machine | 0.9276 | 0.9666 | 0.9054 | 0.8469 | 1.196 |
| Gradient Boosting Classifier | 0.9255 | 0.9660 | 0.9038 | 0.8436 | 5.562 |
| Ada Boost Classifier | 0.9184 | 0.9634 | 0.8954 | 0.8294 | 2.066 |
| Logistic Regression | 0.9174 | 0.9605 | 0.8950 | 0.8284 | 1.077 |
| SVM – Linear Kernel | 0.9157 | 0.0000 | 0.8921 | 0.8243 | 1.014 |
| Ridge Classifier | 0.9114 | 0.0000 | 0.8896 | 0.8193 | 0.967 |
| Linear Discriminant Analysis | 0.9114 | 0.9604 | 0.8896 | 0.8193 | 1.010 |
| K-Neighbors Classifier | 0.9107 | 0.9463 | 0.8870 | 0.8151 | 1.841 |
| Decision Tree Classifier | 0.8754 | 0.8693 | 0.8372 | 0.7364 | 1.175 |
| Quadratic Discriminant Analysis | 0.8574 | 0.9479 | 0.8357 | 0.7307 | 0.972 |
| Naïve Bayes | 0.8392 | 0.9341 | 0.8190 | 0.7034 | 0.922 |

*Note*: Trained with fifteen selected features for components labeled for processors "0" through "5".

Considering the possible consequences of vital data loss, maximizing accuracy and F1 scores while minimizing false negatives outweighs processing time. Thus, all fifteen columns were used when training the final algorithm for enterprise applications if no further data manipulation occurs. PyCaret uses normalization and stratified k-fold cross-validation without hyperparameter tuning to achieve these results.

Table 2. PyCaret Model Comparison 2 Results

| Model | Accuracy | AUC | F1 | MCC | TT (Sec) |
|---|---|---|---|---|---|
| Gradient Boosting Classifier | 0.8589 | 0.9158 | 0.8326 | 0.7221 | 1.300 |
| Light Gradient Boosting Machine | 0.8585 | 0.9138 | 0.8316 | 0.7204 | 0.225 |
| Ada Boost Classifier | 0.8531 | 0.9132 | 0.8287 | 0.7158 | 0.586 |
| K Neighbors Classifier | 0.8421 | 0.8860 | 0.8088 | 0.6813 | 0.361 |
| Random Forest Classifier | 0.8304 | 0.8923 | 0.7876 | 0.6487 | 2.056 |
| Extra Trees Classifier | 0.8253 | 0.8653 | 0.7782 | 0.6352 | 2.558 |
| Logistic Regression | 0.8163 | 0.8727 | 0.7805 | 0.6311 | 0.144 |
| Decision Trees Classifier | 0.8098 | 0.8128 | 0.7548 | 0.5998 | 0.112 |
| SVM – Linear Kernel | 0.8074 | 0.0000 | 0.7906 | 0.6530 | 0.109 |
| Ridge Classifier | 0.7992 | 0.0000 | 0.7457 | 0.5812 | 0.114 |
| Linear Discriminant Analysis | 0.7992 | 0.8763 | 0.7457 | 0.5812 | 0.086 |
| Naïve Bayes | 0.7389 | 0.8551 | 0.6212 | 0.4316 | 0.073 |
| Quadratic Discriminant Analysis | 0.7162 | 0.8373 | 0.5804 | 0.3788 | 0.076 |

*Note*: Models trained using three features from the processor labeled "0": CPU load, temperature, and clock speed.

Individually trained models and their respective Receiver Operator Characteristic (ROC) curves are seen in Figure 4. While each achieves an Area Under the Curve (AUC) of 0.96, Light Gradient Boosting Machine can be seen to have a slightly sharper knee leading to a higher true positive rate for other models' corresponding false positive rates. Random Forest and Extra Trees models both perform poorly with less features.

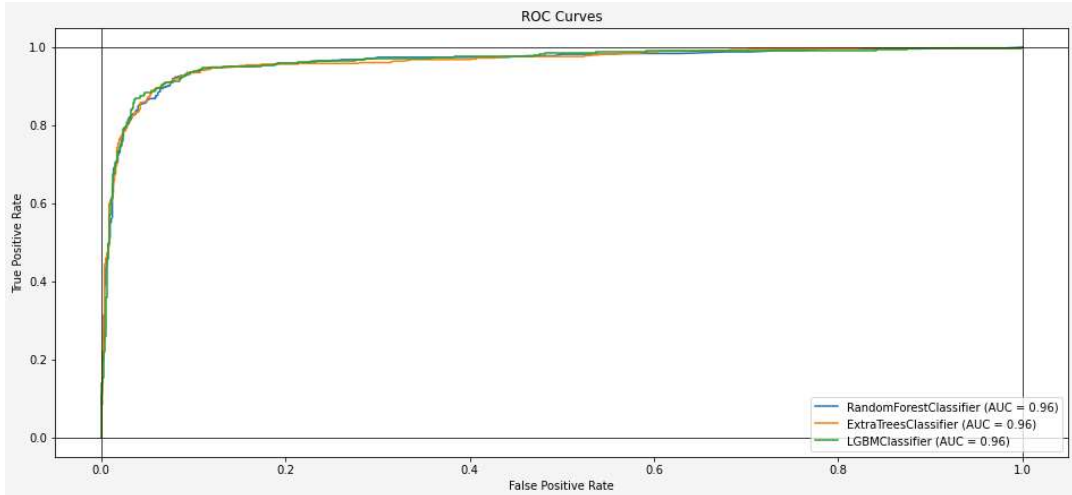

*Figure 4. Receiver Operator Characteristic (ROC) curves with AUC of 0.96*

### 3.2 Dense Neural Network (DNN)

Using TensorFlow's Keras (version 2.4.3), several dense sequential networks are trained and scored. The average performance in all metrics was around 91%. Variance is controlled by a dropout of 0.3 with no other regularization implemented. Test scores tend to be erratic in the earlier training rounds, but stability is eventually reached with low variance (overfitting). False negatives are greatly reduced in these initial training epochs, but plateau and continue to be constant as training persists. The best performing model – trained for 500 epochs – scores 92% accuracy with 134 false negatives (approximately 5%) on unseen test data taking 1.25 hours for training and testing. This method had the lowest F1 score of only 91%; Figure 5 shows the progression of false negative and high variance reduction between train and test sets.

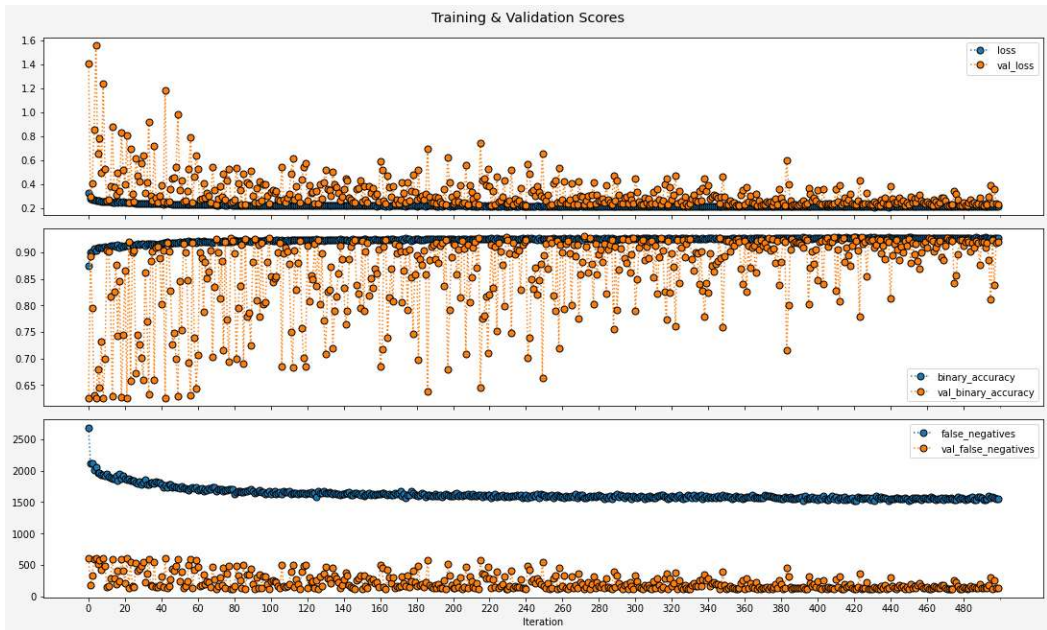

Figure 5. DNN false negatives plateau during training but display less deviation in testing after 500 epochs

### 3.3 Multilayer Perceptron

Data preparation for a multilayer perceptron was performed similarly to preparation for other models, but included additional data representative of zipping operations, to be contrasted with malicious action. In addition, supplementary features were used for this model, specifically RAM load and data values, and all features were normalized to a range between 0 and 1.

Given unbalanced classes, random sampling and under-sampling are used to produce a balanced train/test dataset reflective of possible scenarios. Several MLP models were produced and tested, with the most successful having three hidden layers, trained for 1000 epochs using the Adam optimizer and an α of $1e^{-3}$. This model achieves a 97% accuracy on its test data, as well as 97% precision, recall, and F1 scores. Figures 6–8 display the confusion matrices for Random Forest, Extra Trees, and Light Gradient Boosting Machine, respectively.

### 3.4 Random Forest Classifier

This legacy model achieves results above 90% in all categories using "Gini" criterion and 800 estimators. There are six more false positives than false negatives but varying the random state changes this slightly. While encouraging, the random state is not a viable hyper-parameter, and is set to 42 for reproducibility. The best ccp_alpha was found to be 0.0 with no max depth imposed on the trees. Unlike the ExtraTrees Classifier discussed in the next section, RF uses the optimum point to split the nodes in bootstrapped replicas, reducing high variance.

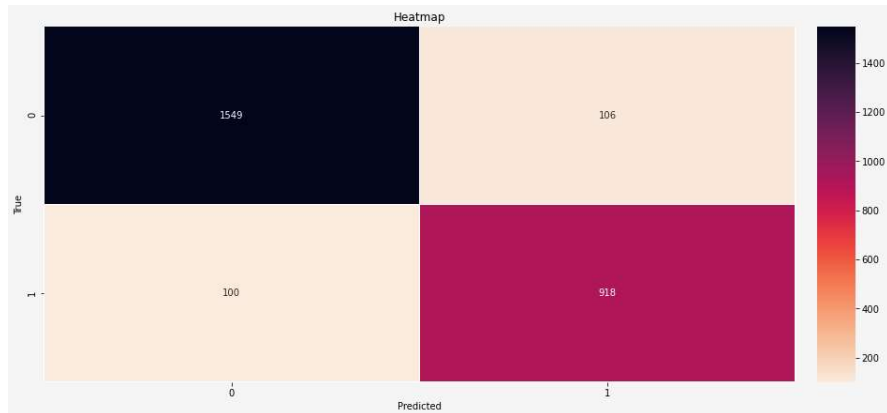

Figure 6. Random Forest confusion matrix with 100 false negatives (3.7%)

### 3.5 ExtraTrees Classifier

While RF uses bootstrapped replicas, ExtraTrees (ET) uses the entire sample when bootstrap is set to false [8]. Using a top-down procedure it builds an ensemble of unpruned regression trees and splits nodes by "choosing cut-points fully at random" [9]. This adds an element of uncertainty to the model while achieving optimization and helps to inherently balance bias and variance. The application of the same method on the entire sample makes ET calculations slightly faster than with RF. When under a ransomware attack, this timing edge could be the deciding factor in detecting and defeating the intrusion.

The model performs somewhat better than RF, achieving 93% inaccuracy and F1, but with fewer false negatives and an equal number (106) of false positives. In contrast, the best number of estimators – 4500 – is over four times that used in the RF model. The best criterion is again "Gini" with no max depth imposed.

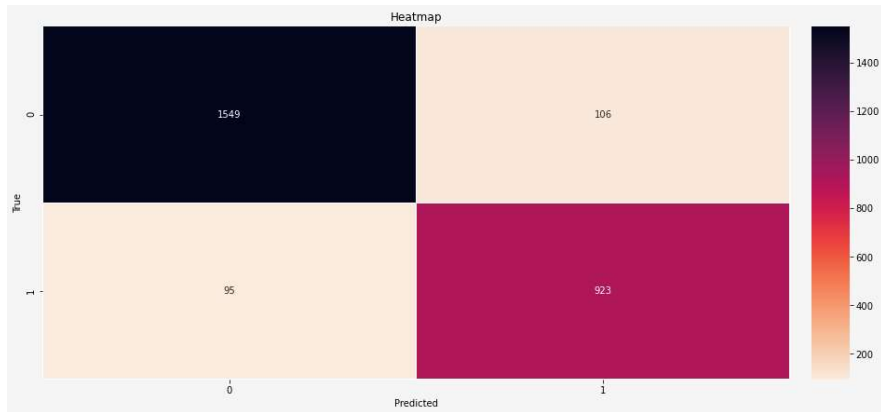

Figure 7. Extra Trees confusion matrix with 96 false negatives (3.6%)

### 3.6 Light Gradient Boosting Machine (LGBM)

The prevalence of tree methods continues with the scores of LGBM achieving the same as RF but with three more false negatives and one less false positive. A grid search found the best boosting_type to be "dart" with 1000 estimators and a learning rate of 0.1. Additionally, the number of leaves is 8, and reg_alpha and reg_lambda are set to 0.0005 and 0.05, respectively.

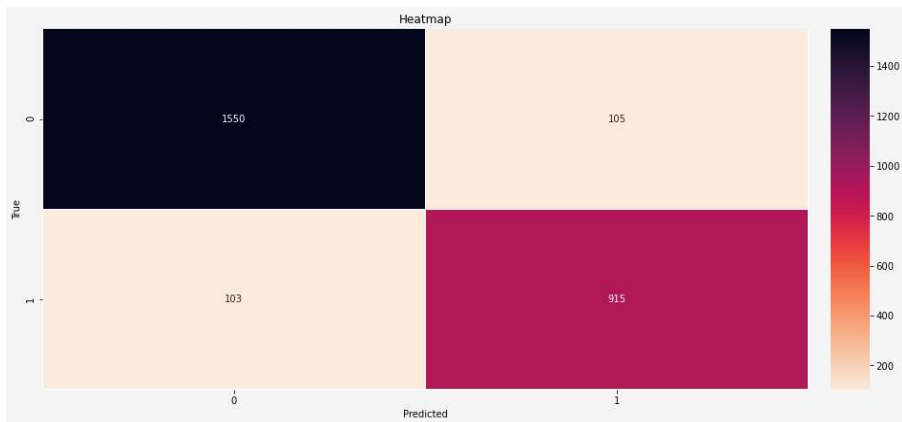

Figure 8. LGBM Classifier confusion matrix with 103 false positives (3.9%)

LGBM was developed to deal with sparse matrices having mutual exclusivity among datasets with a very large number of features [9]. Since there are no missing values and only 15 variables used in this test, LGBM might outperform its tree-based relatives in a more robust and sparser feature space. The bundling strategies that make this method unique do not seem to offer any extra performance enhancement, as seen in Table 3 below.

Table 3. Model Optimization Results

| Model | Accuracy | Precision | Recall | F1-score | MCC | Support |
|---|---|---|---|---|---|---|
| Multilayer Perceptron | 0.97 | 0.97 | 0.97 | 0.97 | 0.83 | - |
| Random Forest Classifier | 0.92 | 0.92 | 0.92 | 0.92 | 0.84 | 2673 |
| Extra Trees Classifier | 0.93 | 0.92 | 0.93 | 0.93 | 0.84 | 2673 |

| Light Gradient Boosting Machine | 0.92 | 0.92 | 0.92 | 0.92 | 0.84 | 2673 |
| Dense Neural Network | 0.92 | 0.92 | 0.91 | 0.91 | 0.83 | 2673 |

*Note:* See the section on Multilayer Perceptron for additional data processing used

## 4. CONCLUSIONS

Scikit-learn's Multilayer Perceptron model achieves the best scores of 97% accuracy and F1 after thorough data preparation. In a PyCaret general comparison of 13 viable ML algorithms with only normalization and stratified cross-validation, the top three performing models are Random Forest Classifier, Extra Trees Classifier, and Light Gradient Boosting Machine. Each method produces accuracy and F1 scores above 90%, supporting the hypothesis that ransomware activity can be detected by physical changes the computer undergoes during file encryption. Constant monitoring of file activity and entropy is not required with this approach. Thus, power and memory usage are significantly reduced. Since this novel method does not rely on known signatures or virtual environments, zero-day attacks are also much more likely to be detected upon deployment.

## ACKNOWLEDGEMENTS

The authors would like to thank the PeopleTec, Inc. Technical Fellows program for encouraging and assisting this research.

## Authors

Erik Larsen, M.S. is a senior data scientist with research experience in quantum physics and deep learning. He completed both M.S. and B.S. in Physics at the University of North Texas, and a B.S. in Professional Aeronautics from Embry-Riddle Aeronautical University while serving as an aviator in the U.S. Army.

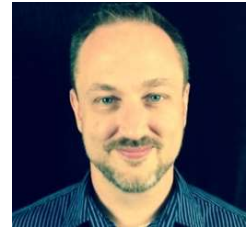

David Noever, Ph.D. has research experience with NASA and the Department of Defence in machine learning and data mining. Dr. Noever has published over 100 conference papers and refereed journal publications. He earned his Ph.D. in Theoretical Physics from Oxford University, as a Rhodes Scholar.

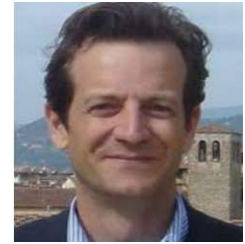

Korey MacVittie, M.S. is a data scientist specializing in machine learning. Prior research includes identifying undervalued players in sports drafting. He completed his M.S. at Southern Methodist University, and a B.S. in Computer Science, and a B.S. in Philosophy from the University of Wisconsin Green Bay.

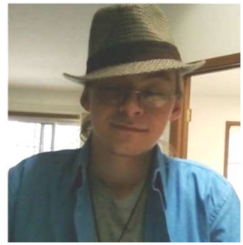